\documentclass[journal]{IEEEtran}
\usepackage{multirow}
\usepackage{cite}
\usepackage{balance}
\usepackage{color}
\usepackage[ruled]{algorithm2e}
\usepackage{epsfig}
\usepackage{epstopdf}
\usepackage{graphicx}

\usepackage{amssymb,amsmath,amsthm}

\usepackage{bm}
\usepackage{pict2e}

\newcommand{\comment}[1]{}

\usepackage{colortbl}

\ifCLASSINFOpdf
\else
\fi
\title{Cloud-based Deep Learning of Big EEG Data for Epileptic Seizure Prediction}
\author{{\bf Mohammad-Parsa Hosseini$^{1}$, Hamid Soltanian-Zadeh$^{2,3}$, Kost Elisevich$^{4,5}$, and Dario Pompili$^{1}$}\\

$^{1}$Dept. of Electrical and Computer Engineering, Rutgers University--New Brunswick, NJ, USA\\

$^{2}$Image Analysis Lab, Dept. of Radiology and Research Administration, Henry Ford Health System, MI, USA\\

$^{4}$Dept. of Clinical Neuroscience, Spectrum Health System, MI, USA\\ 

$^{3}$Division of Neurosurgery, College of Human Medicine, Michigan State University, MI, USA\\
{\tt\small$^{1}$\{parsa,  pompili\}@cac.rutgers.edu, $^{3}$hsoltan1@hfhs.org, $^{4}$konstantin.Elisevich@spectrumhealth.org}
}  

\begin{document}
\maketitle

\markboth{IEEE Global Conference on Signal and Information Processing (GlobalSIP), Greater Washington, DC, Dec 7-9, 2016.}
\maketitle
 
\begin{abstract}
Developing a Brain-Computer Interface~(BCI) for seizure prediction can help epileptic patients have a better quality of life. However, there are many difficulties and challenges in developing such a system as a real-life support for patients. Because of the nonstationary nature of EEG signals, normal and seizure patterns vary across different patients. Thus, finding a group of manually extracted features for the prediction task is not practical. Moreover, when using implanted electrodes for brain recording massive amounts of data are produced. This big data calls for the need for safe storage and high computational resources for real-time processing. To address these challenges, a cloud-based BCI system for the analysis of this big EEG data is presented. First, a dimensionality-reduction technique is developed to increase classification accuracy as well as to decrease the communication bandwidth and computation time. Second, following a deep-learning approach, a stacked autoencoder is trained in two steps for unsupervised feature extraction and classification. Third, a cloud-computing solution is proposed for real-time analysis of big EEG data. The results on a benchmark clinical dataset illustrate the superiority of the proposed patient-specific BCI as an alternative method and its expected usefulness in real-life support of epilepsy patients.
\end{abstract}
\section{INTRODUCTION}
\textbf{Motivation: }Almost one percent of the world's population suffers from epilepsy~\cite{site3}, a chronic disorder characterized by the occurrence of spontaneous seizures. Although the symptoms of a seizure affect any part of the body, the electrical events that produce the symptoms occur in the brain. For about 30 percent of the patients, medications are not curative and, even after surgery, many patients may have spontaneous seizures~\cite{hosseini2014statistical}. Anxiety due to the possibility of a seizure occurring may affect the quality of life of the patients as well as their safety, relationships, work condition, driving and so much more.  

\textbf{Vision: }The use of computers to help physicians in the acquisition, management, storage, and reporting of the biomedical signals is well established~\cite{hosseini2012detection, hosseini2013three,hosseini2012designing,hosseini2011computer}. To this end, Brain-Computer Interfaces~(BCIs) use Electroencephalogram (EEG) which is a measure of brain waves. For example, a BCI system for seizure forecasting can help epileptic patients have a better quality of life. In order for such a BCI system to work effectively, computational algorithms must reliably identify periods of increased probability of seizure occurrence. If the occurrence of seizures could be identified, designing devices to warn patients would be possible and patients could avoid dangerous activities like driving or swimming. Also, medications could be used only when needed to prevent impending seizures.

\textbf{Challenges: }There are many difficulties and challenges in developing a seizure-prediction system as a real-life support for epileptic patients. The first challenge is due to the fact that EEG is not a stationary signal. Therefore, normal and seizure patterns may vary across different patients. As a result, finding a group of manually-extracted features might not scale well to new patterns of seizure activity, and supervised feature extraction may not be sufficient for learning algorithms. The second challenge relates to situations of electrodes implanted within the head that provide for intracranial electroencephalography~(iEEG). This method of brain-signal recording has potential advantages like high spatio-temporal resolution and electro-optic mapping of the dynamic neuronal activity. However, implanted electrodes generate massive amounts of real-time data leading to the big data problem. This situation calls for a safe storage to save the large volume of data and for high computational resources to process the data in real time. 

\textbf{Our Approach: }Signal processing, machine learning, and brain-state prediction need to be carried out in big data in order to develop a practical BCI. The next generation BCI systems may be connected to high-performance computing servers to process medical big data in real-time. Cloud computing is a new Information and Communications Technology (ICT) which enables ubiquitous and on-demand access to computational resources through the global Internet. Our approach is to develop new processing and classification methods to be implemented as a cloud-based BCI. 

\textbf{Contributions: }To address existing challenges, we introduce a cloud-based BCI system for big data problem in epilepsy. Moreover, we have developed a deep-learning unsupervised feature-extraction technique for seizure prediction. Specifically, our contributions include the development of the following novel methods:
\begin{itemize}
  \item A dimension reduction using Principal and Independent Component Analysis to increase the classification accuracy as well as to reduce the computation time and the communication bandwidth. 
  \item A stacked autoencoder as a deep-learning structure to analyze EEG signals for the epileptic seizure prediction. 
   \item A BCI system implemented in the cloud as a safe storage with high computational resources for big data problem generated by implanted electrodes.        
\end{itemize}
The proposed system has the ability of pervasive data-collection and analysis, which is useful in real-life support for epileptic patients. To study the accuracy and performance, the system is evaluated and compared to other methods on a benchmark epilepsy dataset.

\textbf{Paper Outline: }The remainder of this paper is organized as follows. In Sect.~\ref{literature}, we provide a literature review. In Sect.~\ref{solution}, we present our solution including dimensionality reduction, a novel stacked autoencoder as a deep-learning structure to analyze EEG signals, and cloud-computing framework. Then, in Sect.~\ref{results}, we discuss the proof-of-concept prototype of the proposed BCI seizure predictor and show preliminary results. Finally, in Sect.~\ref{conclusion}, the paper is concluded.

\section{Literature Review}\label{literature}
In this section, we provide an overview of the previous studies on seizure prediction systems and big data management of epilepsy. In~\cite{wang2015extracting}, extracting EEG features for epileptic seizure prediction is followed by an elimination-based feature selection method to improve the efficacy and diminish redundant points. In~\cite{brinkmann2015forecasting}, a support vector machine (SVM) algorithm was developed to identify preictal states in continuous iEEG recordings of dogs with naturally occurring epilepsy. In~\cite{zhang2016low}, spectral power and ratios of spectral power extracted from iEEG and processed by a second-order Kalman filter and then input to a linear SVM classifier for epileptic seizure prediction. In ~\cite{lin2016classification}, for classification of preictal and interictal stages, artifact-free preictal and interictal EEG epochs were acquired and characterized with global feature descriptors. In general, existing works have focused on local processing and storage without considering multiple channels and big patient data. In~\cite{parsa}, our group developed a multi-tier distributed computing structure based on Mobile Device Cloud (MDC) and cloud computing for real-time epileptic seizure detection. In this work, we have developed a deep learning structure in the cloud to address the big data analysis problem in epilepsy. In contrast to the existing methods, the proposed method extracts unsupervised features from iEEG patterns to predict seizures.   

\section{Proposed Work}\label{solution}
We propose a seizure prediction system for real-time big data analysis of EEG that can be implemented as a cloud-based service. In Sect.III-A, the data dimension is reduced by principal and independent component analysis. Decreasing the data dimensions decreases the telecommunication bandwidth needed for sending the data to the cloud, increases the classification accuracy by eliminating noise information, and reduces the computational time and energy. In Sect.III-B, a deep learning technique is developed using Stack Autoencoder for unsupervised feature extraction from big unlabeled data. In the end, a softmax layer classifies interictal (baseline) patterns of Preictal (prior to seizure) signals. In Sect.III-C, a cloud based architecture is described for the BCI system.

\textbf{A. Dimensionality Reduction: }
For an efficient analysis of a complex data set, dimensionality reduction is critical. Given a data space $\mathbf{d}\in \mathbb{R}^{N}$, dimension reduction methods~\cite{babagholami2013pssdl,rahmani2016subspace} find a mapping $\mathbf{x}=f(\mathbf{d}):\mathbb{R}^{D}\rightarrow\mathbb{R}^{M}$ ($M<D$) such that the transformed data vector $\mathbf{x}\in \mathbb{R}^{M}$ preserves most of the information of $\mathbf{d}$. In~\cite{parsa}, we proposed a method based on Infinite Independent Component Analysis (I-ICA)~\cite{knowles2007infinite} for the EEG feature selection task. In this paper, to enhance the dimensionality reduction process, a Principal Component Analysis (PCA) method is applied before I-ICA. PCA generates a diagonal covariance matrix from the input data~\cite{minaee2015screen, rahmani2015randomized,joneidi2016union}. Then, using a transformation each dimension is normalized such that the covariance matrix is equal to the identity matrix~\cite{minaee2015fingerprint}. As a result, small trailing eigenvalues are discarded and also computational complexity is decreased by minimizing pairwise dependencies. In this combination, PCA decorrelates the input EEG raw data and the remaining higher-order dependencies are separated by I-ICA. The proposed method for dimensionality reduction is described in Algorithm~\ref{algo:dim}.

\begin{algorithm}[t!]
{
\caption{Dimension Reduction by PCA + I-ICA} 

\label{algo:dim}
\KwIn{D-dimensional EEG raw data $\mathbf{d}=[d_{1},\cdots,d_{D}]^T$}
\KwOut{\ M-dimensional signal $\mathbf{x}=[x_{1},\cdots,x_{M}]^T,$}
$\odot$ denotes the element wise multiplication


\textbf{Compute} Covariance matrix of D

\textbf{Choose} P largest eigenvalues

\textbf{Find} $Y=W^{T}D$ \text{, Y is N dimensional}

\For {$i:=1 \to N$}{
${[Y,X,Z,E]}=concatenation \{\mathbf{y}_i\,\mathbf{x}_i\,\mathbf{z}_i\,\mathbf{e}_i\}_{i=1}^{N}$
}
\For {$k:=1 \to K$}{

\textbf{Define} 
$z_{ki}$ as activity of $k^{th}$ source for $i^{th}$ sample 

\textbf{Define}
$m_{k}=\sum_{i=1}^{N}z_{ki}$ as the active sources

\textbf{Calculate} $p(\mathbf{Z}|\pi_1,...,\pi_K)$ by $\prod_{k=1}^{K}\pi _{k}^{m_{k}}(1-\pi _{k})^{N-m_{k}}$

\For {$i:=1 \to N$}{

\textbf{Define }$\mathbf{g}_{k}$ as the $k^{th}$ column of $\mathbf{G}$

\textbf{Find}
$\mu_{\pm }$ \text{by}
$\frac{g_{k}^{T}e^{\circ}_{ki}\pm \sigma _{e}^{2}}{g_{k}^{T}g_{k}}$

\textbf{Find}
$\sigma ^{2}$ \text{by} 
$\frac{\sigma  _{e}^{2}}{{g_{k}^{T}g_{k}}}$

\textbf{Find} 
$e^{\circ}_{ki}$ \text{by} 
$(e_{ki}|z_{ki}=0)$

\textbf{Calculate} $\mathbf{p}(x_{ki}|\mathbf{G},x_{-ki},y_{i},z_{i})$
\text{by}

\textbf{    if $x_{ki}> 0$}\textbf{ then}
$\mathcal{N}(x_{ki};\mu _{-},\sigma ^{2})$

\textbf{    if $x_{ki}< 0$}\textbf{ then}
$\mathcal{N}(x_{ki};\mu _{+},\sigma ^{2})$
}
}
\textbf{Find} $X$ \text{by}
$\mathbf{G}[\mathbf{X}\odot\mathbf{Z}]+\mathbf{E}$

\textbf{Extract} $\mathbf{x}=[x_{1},\cdots,x_{M}]^T$ from $X$\\
}
\normalsize
\end{algorithm}\
\textbf{B. Deep Network: }A stacked autoencoder which is a class of deep neural networks~\cite{chen2014deep} with two sparse encoders as hidden layers is developed. Stacked autoencoder captures the hierarchical grouping of the EEG input for seizure prediction task. The encoder maps the input to a hidden representation. The size of the second hidden layer is designed less than the first hidden layer so the second encoder learns an even smaller representation of the input data. The deep network structure is shown in Fig 1. Hidden layers are trained individually in an unsupervised method. The training data without labels are used to replicate the input from the output in the training step. To enforce a constraint on the sparsity of the output from the hidden layer, the impact of a sparsity regularizer is controlled. The first autoencoder tends to learn first-order features in the raw EEG input. Using the primary features as the input to second hidden layer, the second-order features are extracted. Then, a softmax layer is trained and the layers are joined to form a deep network. Finally, the deep network is trained one final time in a supervised manner. The pseudocode of the proposed classification method is shown in Algorithm~\ref{algo:Deep}.

The main property of stacked autoencoder is the ability of feature extraction from a large amount of unlabeled data which makes it a suitable solution for the big data problem. A nonlinear transformation is applied to each layer's input and a representation is provided in the output. Thus, there is no need to extract EEG features by hand-engineering techniques for each patient. In deep architecture, multiple nonlinear transformation layers are stacked together to represent a nonlinear function of EEG data. A gradient-log-normalizer of the categorical probability distribution as softmax layer~\cite{guo2016deep} is used to classify the nonlinear function of EEG as interictal or preictal signal in the last layer.
\begin{equation}\label{eq:softmax}
P(c_{r}|\mathbf{x})=\tfrac{P(c_{r})P(\mathbf{x}|c_{r})}{\sum_{k=1}^{K} P(c_{k})P(\mathbf{x}|c_{k})}=\tfrac{exp(a_{r})}{\sum_{k=1}^{K} exp(a_{k})}
\end{equation}
where $a_{k}=ln(P(c_{k})P(\mathbf{x}|c_{k}))$, $P(c_{k})$  is the class prior probability, and $P(\mathbf{x}|c_{k})$ is the conditional probability of the sample given class k.


\textbf{C. Cloud Computing: }Cloud computing provides a “limitless” scale of computing power that can be made available on demand and by way of the Internet makes it ubiquitously available for an extensive global reach. There are many cloud platforms including Microsoft, Google and Amazon AWS. But for the purposes of our study and based on proven use-cases for large scale processing, we will base our reference of cloud usage to the Amazon Cloud, otherwise called Amazon Web Services (AWS). The cloud is generally broken into three layers based on the service provided: (1) Infrastructure as a Service (IaaS); (2) Platform as a Service (PaaS); and (3) Software as a Service (SaaS). These 3 layers will all lend to the different infrastructural setup of the BCI as follows: 

\textit{IaaS} provides computing power, networking, storage and virtual orchestrators and operating systems. It is available at large scale and on demand with the ability to deliver High Performance Computing (HPC) which lends itself well to the processing required with rapid real-time epilepsy monitoring. An applicable BCI system dealing with large amounts of data from distributed electrodes requires storage capability and both rapid and timely event-related mining to produce intelligence in the forms of trends, predictions and recommendations.  With a low cost of entry and ease of setup, the core engine of the BCI can be effectively deployed using the AWS HPC. High Performance Computing processors allow the BCI system to function above a teraflop capacity or 1012 floating-point operations per second allowing for realtime results inspite of large data entry. The Health Insurance Portability and Accountability Act (HIPAA) and its Protected Health Information (PHI) provision also requires service providers to adhere to strict assurrances regarding protection of personal data. A need for encryption and use of AWS HIPAA eligible~\cite{site1} services are required to host the BCI system.
 
\textit{PaaS} uses an open source allowing developers from different constituencies to leverage the BCI to continue developing modules and customized features for their local environment in order to adapt the application to their practices and needs. \textit{SaaS} uses a cloud-based BCI application allowing a good deal of processing power to be made available and distributed globally with decreased reliance on local extensive computer infrastructure in order to complete predictions. Aside from the standard electroencephalographic recording units and other specialized detection tools; run analysis, simulations, and other high-end processes can be initiated from relatively light client applications including smartphone apps.

\begin{figure}
\centering
\includegraphics[width=0.5\textwidth]{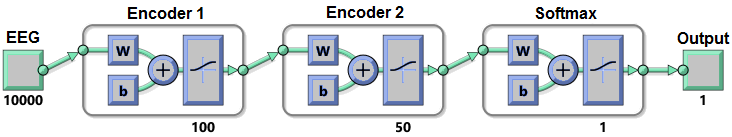}\\
 \vspace{-0.2 cm}
\caption{The encoders from the autoencoders have been used to extract features. To form a deep network, the encoders from the autoencoders are stacked together followed by a softmax layer.}\label{fig:waveletD}
\end{figure}

\begin{algorithm}[t!]
{
\caption{Deep learning by Stacked Autoencoder}\label{algo:Deep}
\KwIn{\ M-dimensional data, $\mathbf{x}=[x_{1},\cdots,x_{M}]^T$}
\KwOut{Classification result as preictal (0) or interictal (1) signal; $output\rightarrow (0,1)$}

\textbf{begin}\\
\For {$i:=1 \to $\#$ Hidden Layers$}{
\textbf{Decrease} the size of the ith hidden layer, P(i)$<$P(i-1)\\
\textbf{Train} unsupervised the ith autoencoder\\
\textbf{Set} explicitly the random number generator seed\\
\textbf{Control} the impact of an L2 regularizer for weights\\
\textbf{Control} the impact of a sparsity regularizer\\
\textbf{Control} the output sparsity from the hidden layer\\
\textbf{Use} the ith feature set for training in the next layer
}
\textbf{Train} supervised a softmax layer to classify ith features\\
\textbf{Stack} the encoders from the autoencoders with softmax\\
\textbf{Compute} the results on the test set $output\rightarrow (0,1)$\\
\textbf{Do} fine tuning by retraining on the training data\\
} \normalsize
\end{algorithm}\

\begin{figure}
\centering
\includegraphics[width=0.5\textwidth]{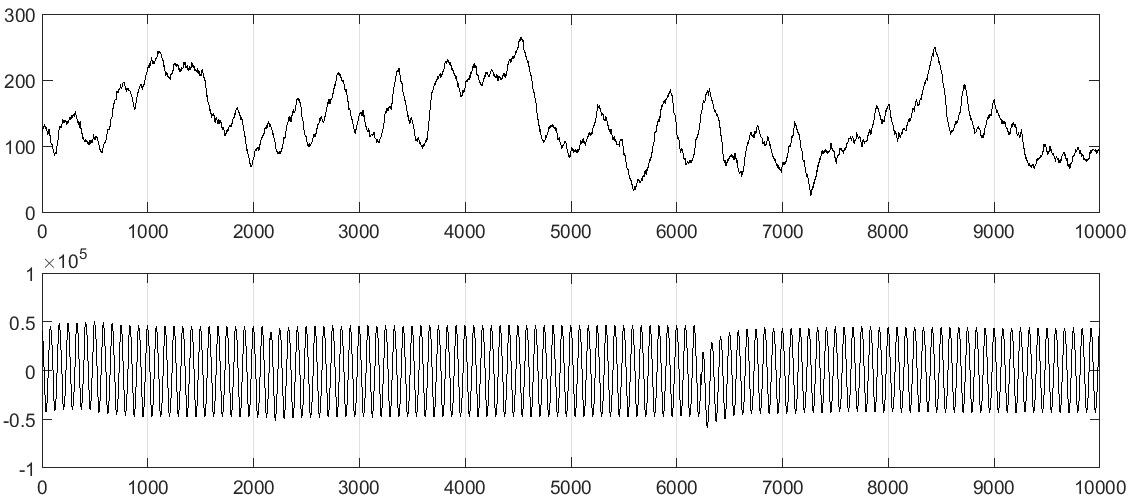}\\
 \vspace{-0.2 cm}
\caption{A comparison of interictal (baseline) iEEG segment on top and preictal (before seizure) iEEG segment on the bottom.}\label{fig:EEG}
\end{figure}

\section{Results}\label{results}
A proof-of-concept prototype of the proposed BCI seizure predictor was developed in the cloud and Autonomic Computing Center (CAC), Rutgers University. In this testbed, we chose to use a benchmark dataset of epilepsy, an HP laptop with intel i5 processor, 8 GB RAM and battery capacity of 4400 mAh, and a supercluster of computers hosted by Amazon Elastic Compute Cloud (EC2). Message Queuing Telemetry Transport
(QMTT) and RESTful Web Service protocols are used for sending data in cloud~\cite{zao2014pervasive}. The clinical iEEG dataset of two epileptic patients (60 interictal and 60 preictal segments) with temporal and extratemporal lobe epilepsy has been used, which was jointly developed by the University of Pennsylvania and the Mayo Clinic, and sponsored by the American Epilepsy Society~\cite{brinkmann2009large}.\footnote{The Dataset recorded by 15 electrodes. Preicatl and interictal data are segmented in 10 minute long clips. The sampling rate is 5000 Hz and the reference recorded voltage is an electrode outside the brain. Preictal data segments covered one hour prior to seizure and seizure horizon is five minutes. The pre-seizure horizon grantees that seizures could be foretasted with enough warning to allow using medications for preventing seizure occurring.} Fig. 2 compares the patterns of the interictal and preictal segments. 

The database consists of a few independent cases with a big data problem. Therefore, algorithms should be regulated against over fitting, and some techniques such as KNN or tree-based algorithms did not work well. However, since the proposed solution extracts the features in an unsupervised manner, the risk of overfitting is decreased. Moreover, to evaluate the generality of the results, we used leave-one-out as an exhaustive cross validation technique~\cite{hosseini2014support, nazem2014,hosseini2015automatic}. Using this technique, the model is fitted to subsets of EEG data and the accuracy of the model is found using the held-out sample~\cite{hosseini2016comparative}.


\begin{table}[h]
\centering
\caption{Confusion matrix. The diagonal elements show the correct decisions} \label{table:one} \scriptsize
 \vspace{-0.2 cm}
\begin{tabular}{lcccccc}
&& \multicolumn{1}{l}{Output interictal} && \multicolumn{1}{l}{Output preictal} && \multicolumn{1}{l}{Total}  \\ \hline
Target interictal   &\vline     & 56 &\vline  & 3  &\vline  & 59     \\
\hline
Target preictal      &\vline    & 4 &\vline & 57  &\vline  & 61   \\
\hline
Total            &\vline     & 60 &\vline & 60  &\vline  & 120   \\

\end{tabular}
\end{table}
 
The confusion matrix of the proposed method is shown in Table~\ref{table:one}. To evaluate the classification ability of the proposed unsupervised feature extraction, the EEG feature sets are used for classification by other methods listed in Table II. The extracted features are based on fast Fourier transform, general energy average, and energy STDV over time for each channel, power spectral density correlation coefficients, partial directed coherence of the coefficients, power in band, low-gamma phase sync, and log of energy in different frequency bands for each channel~\cite{ge2015novel}. Experimental results in Table II show that the proposed deep learning method outperforms previous methods for the EEG seizure prediction task. The feasibility of using cloud computing is analyzed by the network latency offered by Amazon EC2 cloud servers. The Round Trip Time~(RTT) for servers located at different geographical locations (Virginia, Oregon, Singapore, and Ireland) is calculated for 64B EEG segments at 10 days using the “ping” command. The shortest RTT is 15 ms for Virginia server and the longest RTT is 97 ms for Oregon server.

\begin{table}[h]
\centering
\caption{Accuracy, Precision, Sensitivity, Specificity, FPR, and FNR for proposed classification compared with the other methods} \label{table:three} \scriptsize
 \vspace{-0.2 cm}
\begin{tabular}{lcccccc}
\hline
Methods  & \multicolumn{1}{l}{Accuracy} & \multicolumn{1}{l}
{Precision} & \multicolumn{1}{l}{Sensitivity} & \multicolumn{1}{l}{FPR} & \multicolumn{1}{l}{FNR} \\ \hline
Proposed Method            & 0.94   & 0.95   & 0.93   & 0.05   & 0.06   \\
Random Forest              & 0.75   & 0.78   & 0.74   & 0.22   & 0.25   \\
Linear SVM                 & 0.71   & 0.73   & 0.70   & 0.27   & 0.29   \\
Non-linear SVM             & 0.78   & 0.80   & 0.77   & 0.20   & 0.22   \\
MLP Neural Network         & 0.68   & 0.70   & 0.67   & 0.31   & 0.32   \\
\end{tabular}
\end{table}

\balance

\section{Conclusion}\label{conclusion}
Efficiently handling and processing of medical big data can provide useful information about a patient and about diseases. This is now a high-focus area in data science. Intracranially implanted electrodes can be used for seizure prediction preparatory to stimulus delivery for aborting the event. Such electrodes generate considerable amounts of data, calling for safe storage and high computational resources to process big data. On the other hand, iEEG records a larger variety of patterns with fluctuations in amplitude and frequency, making feature extraction a challenging problem. In order to address these two broad issues, we introduced a novel cloud-based BCI to provide real-time seizure prediction from iEEG data. The proposed preprocessing step as a dimensionality reduction provides more accurate classification and decreases energy, computation time, and communication bandwidth. The developed deep-learning methods have the capability for unsupervised feature extraction and, therefore, represent a suitable substitute to manual feature-extraction techniques for classification purposes. These methods extract high-level, complex abstractions for data representations through a hierarchical learning process. The key benefit of the proposed method centers upon the analysis and learning allowed from massive amounts of unsupervised data, making it a practical method for developing a patient-based seizure prediction system. A cloud-based deep-learning method that is able to perform seizure prediction under such circumstances has immediate applicability in the present day.

\section{How to Cite Item}\label{conclusion}
 M.P. Hosseini, H. Soltanian-Zadeh, K. Elisevich, D. Pompili “Cloud-based Deep Learning of Big EEG Data for Epileptic Seizure Prediction,” IEEE Global Conference on Signal and Information Processing (GlobalSIP), IEEE, 2016.

\bibliographystyle{IEEEtran}
\bibliography{lit}
\end{document}